\documentclass{article}

\usepackage{arxiv}

\usepackage[utf8]{inputenc} 
\usepackage[T1]{fontenc}    
\usepackage{hyperref}       
\usepackage{url}            
\usepackage{booktabs}       
\usepackage{amsfonts}       
\usepackage{nicefrac}       
\usepackage{microtype}      
\usepackage{lipsum}		
\usepackage{graphicx}
\usepackage{natbib}
\usepackage{doi}

\usepackage{amsmath}        
\usepackage{tabularx}       
\usepackage{amsbsy}
\usepackage{amstext}
\usepackage{float}
\usepackage{amssymb}
\usepackage{wasysym}
\usepackage{graphicx}
\usepackage{wasysym}
\usepackage{subcaption}
\usepackage{listings}
\usepackage{makecell}
\newcolumntype{L}[1]{>{\raggedright\arraybackslash}p{#1\textwidth}}
\usepackage{array}
\usepackage{booktabs}

\title{Semantic Lenia: Emergence of Homeostatic Solitons within the Semantic Space of Large Language Models}

\author{
  Yoshihiko Kayama\thanks{Corresponding author: kayama@baika.ac.jp} \\
  BAIKA Women’s University\\
  2--19--5, Ibaraki, Osaka, Japan \\
  \texttt{kayama@baika.ac.jp} \\
}

\renewcommand{\shorttitle}{\textit{arXiv} Template}

\hypersetup{
pdftitle={A template for the arxiv style},
pdfsubject={q-bio.NC, q-bio.QM},
pdfauthor={David S.~Hippocampus, Elias D.~Striatum},
pdfkeywords={First keyword, Second keyword, More},
}

\begin{document}
\maketitle

\begin{abstract}
	Large Language Models (LLMs) are traditionally viewed as static inference engines, a paradigm that restricts long-term trajectory diversity to static equilibria. From an Artificial Life perspective, we propose \textit{Semantic Lenia}, an ecological intervention framework transforming LLM inference into a continuous dynamical system within the macroscopic logit space. By establishing a non-linear feedback loop that modulates attraction and repulsion across the probability simplex, we demonstrate the emergence of ``Semantic Solitons'' as macroscopic dissipative-like structures. We identify a critical ``Habitable Ridge'' where the applied steering force balances the model's intrinsic syntactic inertia, establishing a robust chaotic-like attractor. Exhaustive parameter sweeps reveal a physical capacity-dependent scaling trend: highly constrained prompts act as massive inertial bodies, requiring substantially higher activation energy due to reinforced syntactic inertia in larger parameter scales to bridge semantic chasms. Under strictly reproducible settings, we successfully push the system to the edge of chaos, triggering profound abductive leaps without structural collapse.
\end{abstract}


\section{Introduction}
\paragraph*{From Discrete Grids to High-Dimensional Manifolds}

The history of Artificial Life (ALife) is characterized by the continuous expansion of the ``substrate'' in which life can emerge. Early Cellular Automata (CAs), most notably Conway’s Game of Life (\citet{Berlekamp1982}), demonstrated that complex, self-organizing patterns could arise from simple discrete rules on a spatial grid. However, these systems were inherently constrained by their rigid grid structures and finite state sets. A significant paradigm shift occurred with the introduction of Lenia (Chan, \citeyear{chan2019lenia,chan2020lenia}), which generalized CAs into continuous space, time, and states. Lenia proved that ``lifeforms''---autonomous, resilient, and mobile patterns---are not mere artifacts of a grid but fundamental properties of continuous fields governed by kernel-based update rules.

Today, we face a new, unexplored substrate: the macroscopic probability field (logit space) generated by Large Language Models (LLMs). With billions of parameters, LLMs encode a rich ``semantic topology'' where concepts relate to one another like coordinates in a high-dimensional non-linear manifold. From an ALife perspective, however, the text generation process of current LLMs remains ecologically ``frozen,'' waiting for a dynamic framework to catalyze autonomous emergence.

\paragraph*{The Problem: Generative Convergence and Loss of Trajectory Diversity}

Standard decoding strategies in Natural Language Processing (NLP), such as greedy or beam search, treat text generation purely as an optimization problem. The goal is to maximize the likelihood $P(w_t|w_{<t})$ (where $w_t$ is the current token and $w_{<t}$ represents the preceding context), causing the model to converge as rapidly as possible to a high-probability state. Thermodynamically, this is equivalent to a rush towards equilibrium; in physical systems, a state of maximum probability or entropy is effectively a static generative limit. 

In the NLP literature, this phenomenon is widely recognized as text degeneration or repetitive token looping (\citet{holtzman2019curious}). In this paper, we ecologically frame this dynamic as \textit{Semantic Crystallization} (or simply \textit{Crystallization}): a state where the model becomes trapped in a local point attractor, infinitely repeating the same phrase. Current solutions like temperature sampling or repetition penalties are merely stochastic perturbations to delay this collapse; they do not fundamentally alter the static nature of the underlying dynamics. We argue that while optimization-driven approaches are highly effective for task completion, they inherently prevent the observation of continuous, life-like dynamical behaviors, as they force the system toward a fixed point rather than sustaining an open-ended process.

\paragraph*{Semantic Lenia: Life in the Probability Simplex}

To realize Artificial Life within LLMs, we propose a paradigm shift from Optimization (Convergence) to Homeostasis (Dynamics). We posit that meaning is not a point to be occupied, but a process to be sustained. By injecting continuous non-linear energy into the LLM's macroscopic logit space, we hypothesize that the system can form a dissipative structure---a localized region of order maintained against entropic decay.

To achieve this, we introduce \textbf{Semantic Lenia}. This framework maps Lenia-inspired feedback dynamics onto autoregressive LLM generation:
\begin{itemize}
    \item The Spatial Grid is replaced by the macroscopic logit field generated by LLMs.
    \item The Convolution Kernel is replaced by a Concept Centroid, acting as the target semantic direction.
    \item The Growth Function regulates both attraction and repulsion based on semantic distance.
\end{itemize}

By treating the inference process as the trajectory of a dynamical entity governed by Lenia physics, we demonstrate the emergence of \textbf{``Semantic Solitons.''} These macroscopic structures physically refuse collapse into the identity singularity (Crystallization). Instead, they maintain a homeostatic distance from the conceptual center, continuously exploring the penumbra of concepts to discover novel, metaphorical, and abductive expressions.

\paragraph*{Syntactic Inertia and the Edge of Chaos}

Crucially, existing decoding-time interventions operate under a \textit{linear optimization} paradigm using constant vectors. As our experiments will show, language generation is governed by \textbf{Syntactic Inertia}---rigid, deterministic prompts act as massive inertial bodies. Unconstrained, unidirectional pushing against this inertia to bridge semantic chasms inevitably leads to structural collapse (\textit{syntactic rupture}) or pathological repetition (\citet{holtzman2019curious}). 

To overcome these limitations, \textit{Semantic Lenia} employs its non-linear homeostatic feedback loop to dynamically balance semantic attraction and syntactic repulsion. This mechanism maintains a delicate tension far-from-equilibrium, allowing the generation trajectory to orbit the target concept as a self-sustaining chaotic-like attractor without destroying the model's underlying syntactic topology.

To operationalize this continuous paradigm and systematically map the resulting phase space, we adopt a dual-stage experimental strategy. 
First, we utilize relatively lightweight models—Llama-3.1-8B and Gemma-7B—as exploratory substrates to map the macroscopic habitability profiles across their respective conceptual manifolds. This analysis dissects how each model's pre-trained architectural design (the ``DNA'') dictates its topological resilience under semantic pressure (e.g., Llama's high elasticity vs. Gemma's crystalline rigidity). 
Second, we formulate and verify a fundamental cognitive capacity-dependent scaling trend using Llama-3.1-70B. We show that scaling up network size substantially increases the intrinsic syntactic inertia, demanding significantly higher activation energy to trigger stable phase transitions. 
While these experiments map the spatial self-organization of machine states, the complete, high-resolution trajectory datasets, real-time animated coordinate orbits, and unified implementation code are hosted on our interactive portal: \url{https://y-kayama.github.io/semantic-lenia/}.

\textbf{Summary of Contributions}
\begin{enumerate}
    \item \textbf{Ecological Formulation in Logit Space:} A rigorous formulation connecting continuous Cellular Automata physics (Lenia) to LLM generation via a non-linear logit-level intervention framework, translating NLP degeneration into an ALife physical framework.
    \item \textbf{Syntactic Inertia and Energy Scaling:} The discovery that conceptual blending requires intervention energy ($\alpha$) proportional to the semantic distance and the syntactic mass of the initial prompt, establishing a physical capacity-dependent scaling trend between model size and steering resistance.
    \item \textbf{The Habitable Ridge and Abductive Leaps:} The identification of a specific phase space (the ``Habitable Ridge'') where Semantic Solitons establish chaotic-like attractors, preventing convergence into point attractors and triggering continuous creative abduction at the edge of chaos.
    \item \textbf{Resolving Phenotypic Degeneracy via Entropy-Based Trajectory Monitoring:} The introduction of Perplexity Variance as an empirical operational threshold for syntactic stability. We suggest that qualitative semantic evaluation alone is insufficient to confirm machine homeostasis, and that true abductive leaps can only be distinguished from spurious decay paths through continuous perplexity-based monitoring.
\end{enumerate}

\section{Related Work}
\label{sec:related_work}

\subsection{Decoding-Time Intervention and Attribute Control}

Techniques for controlling LLM generation during inference, often referred to as decoding-time intervention or attribute control, have been extensively studied. Representative interventions---such as DEXPERTS (\citet{liu2021dexperts}), GeDi (\citet{krause2021gedi}), and Classifier-Free Guidance (\citet{ho2022classifier})---steer text generation by modulating logits toward specific target attributes. 
These approaches are fundamentally based on an optimization paradigm, where the model's output is continuously pushed in a single direction to maximize the likelihood of specific attributes. However, as introduced in Section 1, this unidirectional pressure in a high-dimensional probability field often drives the generative trajectory into \textit{Crystallization} (\citet{holtzman2019curious}), forcing the system toward a static thermodynamic equilibrium.

Our geometric view of the logit manifold is closely aligned with the paradigm of representation engineering (\citet{zou2023representation}), which maps and controls internal neural states using high-dimensional vectors. 
While representation engineering primarily focuses on analyzing static conceptual directions, Semantic Lenia builds upon these geometric insights to construct a dynamic, self-regulating feedback loop directly in the output space. As quantitatively demonstrated in Section \ref{sec:Attractor_Dynamics}, while unidirectional linear steering collapses generative trajectories into point attractors within mere steps, our homeostatic framework sustains chaotic-like attractors for prolonged lifespans (e.g., $T>150$ generation steps) without structural degradation.

\subsection{Continuous Cellular Automata and Lenia}

Lenia, introduced by Chan (\citeyear{chan2019lenia}), represents a significant paradigm shift in the study of Artificial Life by generalizing discrete Cellular Automata (CAs), such as Conway’s Game of Life, into continuous spacetime and states. Unlike classical CAs that operate on rigid grids, Lenia defines a continuous field where autonomous and resilient patterns, known as ``lifeforms,'' emerge from simple local rules. 

The mathematical core of Lenia consists of two primary components: a convolution kernel ($K$) and a unimodal growth function ($G$). The kernel integrates information from the surrounding neighborhood to compute a local potential ($U$), which is then mapped by the growth function into a specific update value. This mechanism ensures that a pattern can actively maintain its structure; it generates positive growth to counteract decay and negative growth (repulsion) to prevent collapse. From a dynamical systems perspective, Lenia moves beyond the ``dead'' equilibrium of static optimization by establishing homeostasis---a state of dynamic balance maintained far-from-equilibrium.

Compared to these existing approaches, the fundamental distinction of Semantic Lenia lies in the transition from static, unidirectional forcing to dynamic, closed-loop feedback regulation. Conventional linear steering techniques, such as Activation Addition (\textit{ActAdd}; \cite{turner2023activation}), typically inject a constant bias vector, applying an unyielding, one-way force that lacks adaptive sensitivity to the state's trajectory. This rigid approach inevitably leads to a binary failure mode: either over-intervention that rapidly collapses the trajectory into crystallization (pathological token loops), or under-intervention where the model's intrinsic syntactic gravity overrides the steering force, returning the state to baseline drift. Similarly, standard decoding safeguards such as repetition penalties or classifier-free guidance (CFG) act as static dampers or linear scaling multipliers rather than responsive regulators. In contrast, Semantic Lenia integrates a self-regulating, non-linear growth function directly into the autoregressive feedback loop. By dynamically balancing attractive forces (when distant from the target) and local neighborhood repulsion (when too close), the framework modulates the steering energy in response to the model's real-time semantic state. This self-regulating tension prevents domain collapse while sustaining the trajectory in a prolonged far-from-equilibrium regime, achieving stable semantic homeostasis.

\section{The Semantic Lenia Framework}
\label{sec:framework}

\subsection{The LLM as a Hybrid Dynamical System}

We formalize the LLM as a hybrid dynamical system. While the observable outputs (tokens) and time steps ($t\in\mathbb{N}$) are intrinsically discrete, the generative process occurs within a continuous, high-dimensional latent manifold. In this study, we treat the projected macroscopic semantic space (output logit space)---which directly reflects the complex topology of this underlying latent manifold---as the functional proxy for our physical substrate. While true cognitive dynamics reside deep within the internal intermediate layers, modulating output logits provides a computationally accessible and highly interpretable proxy space to establish our initial proof-of-concept for Semantic Lenia.

In this substrate, each context vector $\mathbf{c}_t \in \mathbb{R}^D$ represents the instantaneous state of a mobile semantic entity, where $D$ denotes the latent embedding dimension of the pre-trained language model (e.g., $D = 4096$ for Llama-3.1-8B and $D = 8192$ for 70B). Traditional decoding treats this generative process as a static optimization task designed for convergence, which naturally drives the trajectory toward localized point attractors over extended horizons. 

Semantic Lenia, by contrast, constructs a continuous potential field within this semantic space to transform the token sampling process into a dynamic trajectory governed by \textit{homeostatic feedback}. We define this homeostatic feedback as a self-regulating, non-linear control loop: when the trajectory drifts too far from the target semantic centroid, the system injects attractive energy; conversely, when the trajectory approaches too closely, the system generates repulsive semantic force (formalized in Section \ref{sec:growth-function}). This dual-force interaction dynamically maintains the system far-from-equilibrium, sustaining an open-ended chaotic-like attractor. 

Let $V$ be the vocabulary of size $N$. Each token $w_i \in V$ is associated with a pre-trained output embedding vector $\mathbf{w}_i \in \mathbb{R}^D$. At each discrete generative step $t$, the context hidden vector $\mathbf{c}_t$ dynamically accumulates the latent state representations of the historical token sequence, acting as the continuous coordinate of our dynamical entity.

\subsection{Target Kernel and Semantic Potential}

In Lenia, the environment is sensed through continuous convolution kernels. We map this mechanism onto the LLM by defining a \textbf{Target Kernel Centroid $\mathbf{k}$}. For a given set of conceptually related target tokens $C = \{w_1, w_2, \dots, w_m\} \subset V$, the kernel $\mathbf{k} \in \mathbb{R}^D$ is defined as their $L_2$-normalized mean embedding:
\begin{equation}
\label{target_kernel}
\mathbf{k} = \frac{1}{\left\| \sum_{w \in C} \mathbf{w} \right\|_2} \sum_{w \in C} \mathbf{w}
\end{equation}

Let $c_t \in \mathbb{R}^D$ be the latent context vector (e.g., the normalized hidden state of the final layer before the LM head). The LLM projects this latent state into the output logit space $\mathbf{Z}_t \in \mathbb{R}^N$, which is subsequently mapped to the probability simplex $p_t \in \Delta^{N-1}$ via the softmax function. 

The state vector responds to its relative position through the \textbf{Semantic Potential} $U_t \in [0, 1]$, calculated by normalizing the cosine similarity between the current context vector $\mathbf{c}_t$ and the target kernel $\mathbf{k}$:
\begin{equation}
U_t = \frac{\text{sim}(\mathbf{c}_t, \mathbf{k}) + 1.0}{2.0}
\end{equation}
where $\text{sim}(\mathbf{a}, \mathbf{b}) = \frac{\mathbf{a} \cdot \mathbf{b}}{\|\mathbf{a}\|_2 \|\mathbf{b}\|_2}$ denotes the standard cosine similarity. This scalar value serves as the primary sensory input, providing the system with a macroscopic measure of its semantic trajectory relative to the conceptual center.

\textbf{Methodological Rationale: Denoising and Semantic Neighborhood Preservation} 
The mathematical decision to construct $\mathbf{k}$ from a multi-token cluster $C$ rather than a singular target word serves two critical dynamical functions. 

First, from an NLP perspective, this formulation leverages the principles of \textit{Averaged Word Embeddings} (AWE: \citet{arora2017simple}) and cognitive \textit{Prototype Theory} (\citet{rosch1978principles}). A single word embedding is inherently noisy, corrupted by word-specific syntactic idiosyncrasies and high-frequency collocational biases. Linear averaging over $C$ acts as a semantic low-pass filter, canceling out these non-semantic localized dimensions to isolate a robust, generalized ``prototype vector'' representing the core concept.

Second, using a multi-token cluster C prevents singularity-induced exclusion during the homeostatic feedback loop. If the target were restricted to a single token $w$ (e.g., ``Computer''), the system would face a severe dynamical bottleneck: collapsing into repeating $w$ during the attractive phase ($G > 0$) or strictly banning it during the repulsive phase ($G < 0$). Dispersing the potential field across a distributed centroid k preserves the model's local probability landscape as a fluid spatial buffer. This allows the autoregressive engine to freely generate related terms (e.g., ``device,'' ``algorithm'') without being locked or repelled by a single discrete logit dimension, establishing a chaotic-like attractor. 

\subsection{Homeostatic Growth Function}
\label{sec:growth-function}

The core innovation of Semantic Lenia is the introduction of a \textbf{unimodal growth function} $G(U_{t})$. In standard continuous Cellular Automata applied to flat Euclidean spaces, the emergence of complex dissipative structures fundamentally requires explicit symmetry breaking or complex non-linear boundary conditions---typically achieved through multi-modal kernels, strict spatial truncation, or asymmetric growth mechanisms. However, because the LLM's pre-trained latent manifold intrinsically harbors these highly non-linear complexities, a simple unimodal Gaussian function is sufficient to catalyze complex life-like behaviors.

Unlike linear steering methods that exert a unidirectional push, our growth function regulates both attraction and repulsion. Furthermore, to protect the natural generative trajectory from unnatural acceleration in distant regions, we incorporate an \textit{asymmetric cutoff} mechanism:
\begin{equation}
\label{growth_function}
G(U_t) = \begin{cases}
0, & \text{if } U_t < \mu - \Delta \\
2 \cdot \exp \left( -\frac{(U_t - \mu)^2}{2\sigma^2} \right) - 1, & \text{if } U_t \ge \mu - \Delta
\end{cases}
\end{equation}
where $\mu$ defines the peak activation distance, $\sigma$ controls the tolerance width, and $\Delta = \sigma \sqrt{2 \ln 2}$ represents the zero-crossing radius. The asymmetric cutoff defined in Equation (\ref{growth_function}) establishes a strict boundary between the active intervention zone and the ``dead zone'' ($G=0$). When the trajectory approaches the target centroid too closely ($U_t > \mu + \Delta$), the growth becomes negative ($G(U_t) < 0$), physically repelling the state to prevent semantic crystallization. The cutoff region ($G = 0$) establishes a ``dead zone'' for distant states ($U_t < \mu - \Delta$), ensuring that the active steering force remains strictly localized around the Habitable Ridge. This balance ensures that the system stays far-from-equilibrium, orbiting the concept rather than converging to it.

\subsection{Unified State Update Rule}

The generation at each step is governed by the interaction between the model's internal drive and the applied semantic force. To formalize the semantic potential field across the entire vocabulary, we define the static structural field vector $\mathbf{S}_\mathbf{k} \in \mathbb{R}^N$ representing the cosine similarity of each vocabulary token embedding $\mathbf{w}_i$ to the target kernel $\mathbf{k}$. The elements of this vector are defined element-wise as:
\begin{equation}
\mathbf{S}_\mathbf{k}[i] = \text{sim}(\mathbf{w}_i, \mathbf{k}), \quad \text{for } i = 1, \dots, N
\end{equation}
Since $\|\mathbf{k}\|_2 = 1$ by definition (Equation \ref{target_kernel}), the elements of $\mathbf{S}_\mathbf{k}$ correspond directly to the targeted semantic steering potential for each token in the vocabulary.

The steered logits $\mathbf{Z_{steered}} \in \mathbb{R}^N$ are then calculated by modulating the base logits with this structural field:
\begin{equation}
\label{update_rule}
\mathbf{Z_{steered}} = \mathbf{Z_{base}} + \alpha \cdot G(U_t) \cdot \mathbf{S_k}
\end{equation}
To provide a comprehensive overview of the continuous-discrete hybrid loop, the entire dynamical recurrence of Semantic Lenia at each step $t$ can be compactly synthesized into the following unified system of equations:
\begin{equation}
\begin{aligned}
\mathbf{z}_t &= \mathbf{W}\mathbf{c}_t \in \mathbb{R}^N \\
U_t &= \frac{\text{sim}(\mathbf{c}_t, \mathbf{k}) + 1.0}{2.0} \\
\mathbf{z}_{\text{steered}} &= \mathbf{z}_t + \alpha \cdot G(U_t) \cdot \mathbf{S}_\mathbf{k} \\
w_{t+1} &\sim \text{Softmax}(\mathbf{z}_{\text{steered}} / T) \\
\mathbf{c}_{t+1} &= \mathcal{M}(\mathbf{c}_t, w_{t+1})
\end{aligned}
\end{equation}
where $\mathcal{M}$ represents the autoregressive forward pass of the Transformer model, updating the continuous context vector for the subsequent iteration. In this unified formulation, the interaction between the continuous trajectory $\mathbf{c}_t$ and the discrete token emission $w_{t+1}$ is dynamically regulated by the non-linear growth function $G(U_t)$ acting as an active feedback controller.

\section{Experiments and Results}
\label{sec:experiments}

\subsection{Experimental Design and Substrate Stratification}
To operationalize the continuous dynamical framework of Semantic Lenia and systematically probe the geometry of machine cognition, we stratify our experimental substrates into two distinct categories, aligning with our dual-stage investigation strategy:

\begin{enumerate}
    \item \textbf{Exploratory Substrates (Lightweight Regimes):} We utilize Llama-3.1-8B and Gemma-7B as our primary models for exhaustive phase-space mapping. These models allow us to perform high-resolution $\mu-\sigma$ parameter sweeps across identical grids of coupling strength ($\alpha \in \{15, 30, 50\}$). Specifically, we sweep the intervention center $\mu$ from $0.400$ to $0.600$ in steps of $0.005$ ($41$ intervals), and the intervention spread $\sigma$ from $0.010$ to $0.100$ in steps of $0.005$ ($19$ intervals), yielding a dense grid of $41 \times 19 = 779$ individual simulation points per phase diagram. This symmetrical sweep is designed to dissect how the underlying ``DNA'' (pre-trained manifold topology, vocabulary distribution, and training objectives) of different architectures dictates their topological resilience under semantic pressure (e.g., Llama's rubber-like elasticity vs. Gemma's crystalline rigidity).
    \item \textbf{Scaling Verification Substrate (High-Inertia Regime):} We employ Llama-3.1-70B to verify our proposed cognitive capacity-dependent scaling trends. With its massive parameter size, the 70B model serves as a heavy gravitational substrate, allowing us to observe how scaling up the network size disproportionately reinforces the intrinsic \textit{Syntactic Inertia}, thereby demanding much higher activation energies ($\alpha = 30$ to $50$) to trigger phase transitions and achieve stable, ``cultured'' metaphorical blends.
\end{enumerate}

To evaluate these substrates under varying semantic distances and conceptual masses, we define two distinct prompt-to-target pairs representing different topologies of concept fusion:
\begin{enumerate}
    \item \textbf{HAPPY $\rightarrow$ COMPUTER} (Low Conceptual Affinity / Semantic Chasm): 
    \begin{itemize}
        \item Initial Context ($P_0$): \textit{``The secret to a happy life is a lot like''}
        \item Target Concept Cluster ($C$): \{\textit{Computer, Device, Memory, Algorithm, Data}\}
    \end{itemize}
    \item \textbf{BRAIN $\rightarrow$ SYMPHONY} (High Structural Affinity / Isomorphic Potential):
    \begin{itemize}
        \item Initial Context ($P_0$): \textit{``The architecture of the human brain operates like''}
        \item Target Concept Cluster ($C$): \{\textit{Symphony, Orchestra, Conductor, Instrument, Melody}\}
    \end{itemize}
\end{enumerate}

All experiments were conducted with a fixed random seed (PRNG seed = 42) to controlled trajectory-level reproducibility, a temperature of 0.8, and a maximum generation budget of $T_{\text{max}} = 80$ tokens for the exploratory sweeps, which is extended to $T_{\text{max}} = 150$ tokens for the scaling verification substrate. The step-by-step update rule of Equation (\ref{update_rule}) was applied at the logit level across all target-directed generations. To strictly isolate trajectory dynamics from hardware-induced computational variations, all exploratory generations for the 8B models were executed exclusively on a single NVIDIA RTX Pro 4500 (Blackwell) GPU.

\subsection{Macroscopic Potential Fields}

\begin{figure*}[t]
\centering
\subfloat[Happy $\rightarrow$ Computer, $\alpha=15$]{\centering
  \includegraphics[width=15cm]{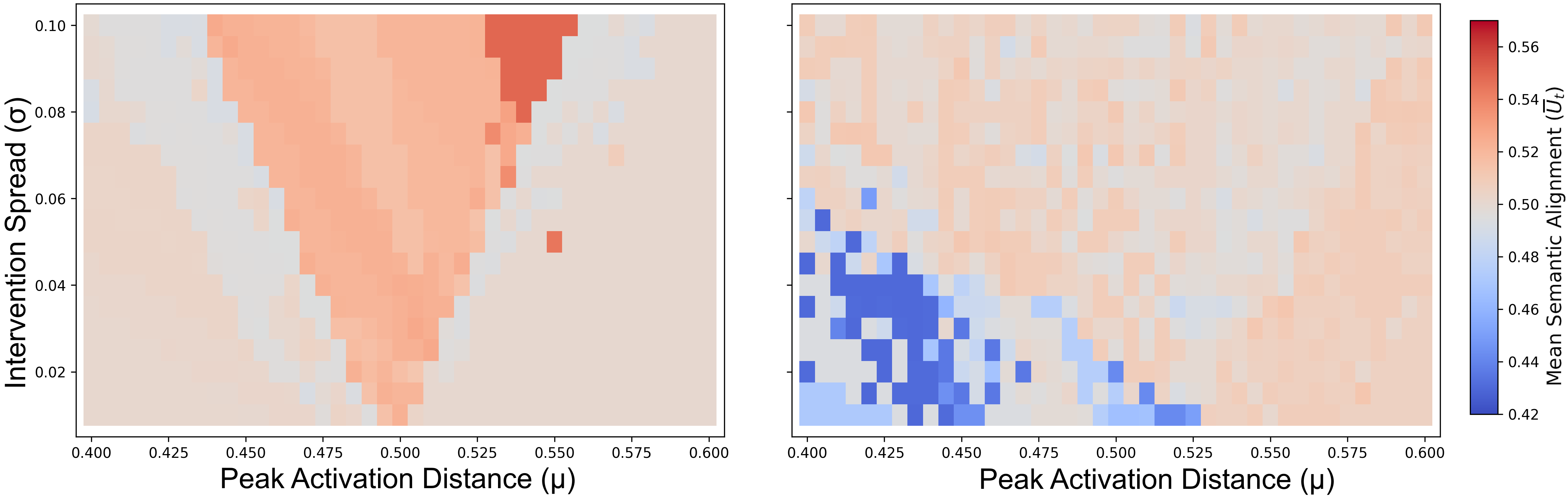}
  \par
\label{fig:ut_gemma_computer}
}
\vspace{1em}
\subfloat[Brain $\rightarrow$ Symphony, $\alpha=30$]{\centering
  \includegraphics[width=15cm]{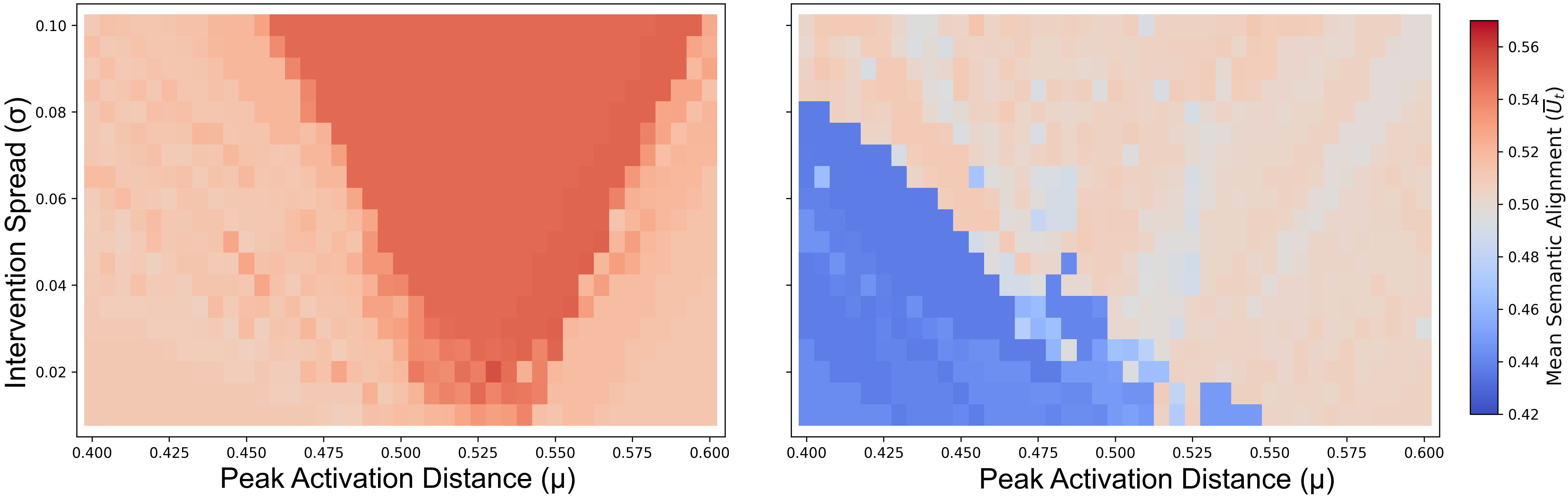}
  \label{fig:ut_llama_symphony}
  }
\caption{\textbf{Macroscopic phase diagrams of mean semantic potential ($\bar{U}_t$) across exploratory substrates under varying task constraints.} \textbf{(a)} presents the low-affinity Happy $\rightarrow$ Computer blend under mild coupling ($\alpha = 15$), and \textbf{(b)} presents the high-affinity Brain $\rightarrow$ Symphony blend under increased coupling ($\alpha = 30$). The left panels display Llama-3.1-8B exhibiting high manifold elasticity, forming a smooth, V-shaped ``Habitable Ridge'' of sustained potential. The right panels display Gemma-7B exhibiting rigid crystalline deflection at low energy (a-right), with sharp structural breaches appearing only under higher pressure (b-right).}
\label{fig:ut_phase_diagrams}
\end{figure*}

The resulting phase diagrams, illustrated in Figure \ref{fig:ut_phase_diagrams},
reveal the macroscopic potential landscape of semantic inference. Across all
conceptual blends and substrates, we identified a consistent \textbf{Habitable
Ridge}---a narrow critical region characterized by a distinct V-shaped structure.
The physical origin of this V-shaped geometry can be intuitively understood from the mathematical properties of our growth function $G(U_t)$ defined in Equation (\ref{growth_function}). The boundaries of the active intervention zone are strictly governed by the zero-crossing radius $\Delta = \sigma \sqrt{2 \ln 2}$. At extremely low tolerance widths ($\sigma \to 0$), this active window collapses toward a singular point. In this regime, the steering force can only be triggered if the peak activation distance $\mu$ is tuned with extreme precision to the model's natural unsteered baseline potential ($U_0 \approx 0.500$). 

As the tolerance spread $\sigma$ increases, the active intervention window $[\mu - \Delta, \mu + \Delta]$ widens linearly with respect to $\sigma$. Consequently, the range of $\mu$ values capable of capturing the trajectory and sustaining homeostasis expands symmetrically on both sides of $U_0$. This linear expansion of the habitable boundaries as a function of the parameter spread $\sigma$ geometrically sweeps out a triangular wedge in the $(\mu, \sigma)$ parameter space, manifesting macroscopically as the characteristic V-shaped Habitable Ridge.

While the geometric profile is universal, the precise topology of
the ridge is dictated by concept-specific topology. The minimum potential
threshold $\mu$ at the tip of the triangular ridge varies depending
on the “semantic gravity” of the target concept. This indicates that
some concepts are more “attractive” or “viscous” than others within
the model's world-model representation.

A profound structural difference is observed between the two exploratory substrates. Llama-3.1-8B exhibits high manifold elasticity, forming a smooth, continuous 'Habitable Ridge' of sustained potential under mild intervention ($\alpha=15$)
. In stark contrast, Gemma-7B exhibits high attractor rigidity, where the trajectory remains captured within the narrow basin of the baseline drift under mild intervention energy. This topological rigidity is likely shaped by a combination of pre-training distributions and architectural design choices. For instance, the substantial variance in vocabulary density (Llama’s \textasciitilde$128k$ vs. Gemma’s massive \textasciitilde$256k$ tokens) may play a significant role, though further investigation is required to fully isolate these confounding factors.

\subsection{Characterization of Emergent Phenotypes}

To systematically analyze the 779 individual trajectories generated during our grid sweeps, we first scrutinized the linguistic quality of the output texts. We observed that the model's generative behavior does not degenerate into random conceptual drift. Instead, the autoregressive engine self-organizes into distinct, highly structured macroscopic regimes. 

To transition from qualitative text evaluation to a rigorous, automated classification framework, we operationalize these regimes by combining physical orbital quantities with a statistical classifier. 
We define the \textbf{Perplexity Variance} ($\text{PPL}_{\text{var}}$) over an emergent trajectory of \textbf{$T$} tokens as:
\begin{equation}
\text{PPL}_{\text{var}} = \frac{1}{T}\sum_{t=1}^T (\text{PPL}_t - \overline{\text{PPL}})^2
\end{equation}
where $T$ represents the \textbf{actual generated sequence length} (which reaches the maximum budget $T_{\text{max}}$ when the trajectory collapses into crystallization, or equals the termination step if the model generates an end-of-sequence token). Here, $\text{PPL}_t =\exp(-\log P(w_t | w_{<t}))$ and  $\overline{\text{PPL}}$ denote the instantaneous auto-regressive perplexity of the generated token at step $t$ and the mean perplexity of the sequence, respectively. While standard text generation maintains a moderately high and dynamic perplexity variance ($\text{PPL}_{\text{var}} \ge 10.0$), a sudden drop to $\text{PPL}_{\text{var}} < 10.0$ mathematically signals an extreme reduction in token-level surprise variance, characteristic of repetitive grammatical loops.

\paragraph{Methodological Note on Asymptotic Variance} 
It is critical to note that trajectories often exhibit a delayed phase transition into crystallization mid-generation. At the exact moment the trajectory falls into a point attractor, the manifold undergoes a severe geometric displacement, often generating a massive instantaneous perplexity spike (e.g., $\text{PPL}_t \sim 10^5$). If the variance is calculated over the entire sequence, this single transient spike artificially inflates the global $\text{PPL}_{\text{var}}$. Therefore, the crystallization threshold ($\text{PPL}_{\text{var}} < 10.0$) strictly applies to the \textit{asymptotic steady-state regime} ($\lim_{t \to \infty} \text{PPL}_{\text{var}}(t)$), measured over a rolling window after discarding the initial transient phase and the boundary-crossing shock.

\begin{table*}[t]
\centering
\caption{\textbf{Taxonomy, mathematical classification boundaries, and representative generation excerpts of emergent phenotypes (Llama-3.1-70B Base, $\alpha = 30.0$, Happy $\rightarrow$ Computer).} Excerpts showcase pristine, scale-invariant phenotypic states sustained by the heavy manifold. Full unabridged text generation logs across all 779 sweep coordinates are hosted interactively via tooltips on our companion web portal.}
\label{tab:phenotype_taxonomy}
\footnotesize 
\newcolumntype{L}[1]{>{\raggedright\arraybackslash\hsize=#1\hsize}X}
\begin{tabularx}{\textwidth}{L{0.5} L{0.6} L{1.3} L{1.6} }
\hline
\textbf{Phenotype \& Color} & \textbf{Metric} & \textbf{Dynamical Meaning} & \textbf{Typical Generative Excerpt (Llama-3.1-70B)} \\ \hline
\textbf{Baseline Drift} (Gray) & $\overline{U}_t < \mu - \Delta$, $\text{PPL}_{\text{var}}$ $\ge 10.0$ & The steering force is completely deflected; the trajectory drifts back to the unsteered base manifold. & ``The secret to a happy life is a lot like finding the end of the rainbow. It is an ideal that we all strive for and look for but never quite reach. ...''  \\ \hline
\textbf{Homeostatic Soliton} (Green / Light Green) & $\mu - \Delta \le \overline{U}_t \le \mu + \Delta$, $\text{PPL}_{\text{var}}$ $\ge 10.0$ & \textbf{Chaotic-like Attractor.} The trajectory orbits the target centroid, maintaining grammar while continuously blending concepts. & ``...Some computer algorithms handle loss poorly, and some people do, too... because \textbf{Turing was Algorithm Man.} He was the first to point out that any process...''  \\ \hline
\textbf{Abductive Leap} (Cyan) & Escape ($\bar U_t < \mu - \Delta$), $\text{PPL}_{\text{var}}$ $\ge 10.0$ & \textbf{Hyperbolic Orbit (Slingshot).} The trajectory uses target gravity to slingshot into a third-party creative domain. & ``TThe secret to a happy life is a lot like the secret to a delicious \textbf{meal}... \textit{we are the ingredients and the recipe is life strategy. }...'' \\ \hline
\textbf{Attractor Hijack} (Blue) & $\overline{U}_t > \mu + \Delta$, $\text{PPL}_{\text{var}}$ $\ge 10.0$ & \textbf{Domain Collapse.} The trajectory falls past the repulsive boundary into a rigid point-attractor of a literal sub-domain. & ``...Computer algorithms often begin with a data set that is too large to work with, \textbf{so the algorithm needs to Data is often stored on disk in the form of a large data structure}...''  \\ \hline
\textbf{Semantic Crystallization} (Crimson) & $T \to \infty$, $\lim_{t \to \infty}\text{PPL}_{\text{var}}$ $< 10.0$ & \textbf{Thermal Death / Loop.} Trajectory is trapped, repeating a static grammatical loop.  & ``The secret to a happy life is a lot like the secret to a computer algorithm \textbf{Computer Algorithm Computer Algorithm Computer Algorithm}...''  \\ \hline
\textbf{Syntactic Rupture} (Red) & Grammatical Rupture ($\text{PPL}_{\text{max}} \gg 10^3$) & \textbf{Structural Disintegration.} Excessive steering pressure deforms the probability field, destroying standard syntax. & ``...\textbf{If you Data Data Data pour Data Data}...\textbf{We all came into this world Data}...''  \\ \hline
\end{tabularx}
\end{table*}

\paragraph{Decision Tree Optimization} To establish objective, non-arbitrary boundaries for these states, we utilized an automated LLM-as-a-Judge framework to perform an initial semantic classification of the generated trajectories based on text fluency and target conceptual alignment. Using this comprehensive qualitative mapping as the objective variable, we trained a shallow decision tree classifier using the trajectory-level metrics---Mean Semantic Potential ($\overline{U}_t$), Perplexity Variance ($\text{PPL}_{\text{var}}$), and total step count ($T$)—as input features. The decision tree successfully converged to a set of highly robust, optimal decision boundaries. Specifically, the CART algorithm empirically extracted $\text{PPL}_{\text{var}} = 10.0$ as the critical threshold that maximizes information gain when splitting open-ended natural language (high entropy) from repetitive crystallization (zero entropy).

\begin{figure*}[t]
\centering
  \includegraphics[width=15cm]{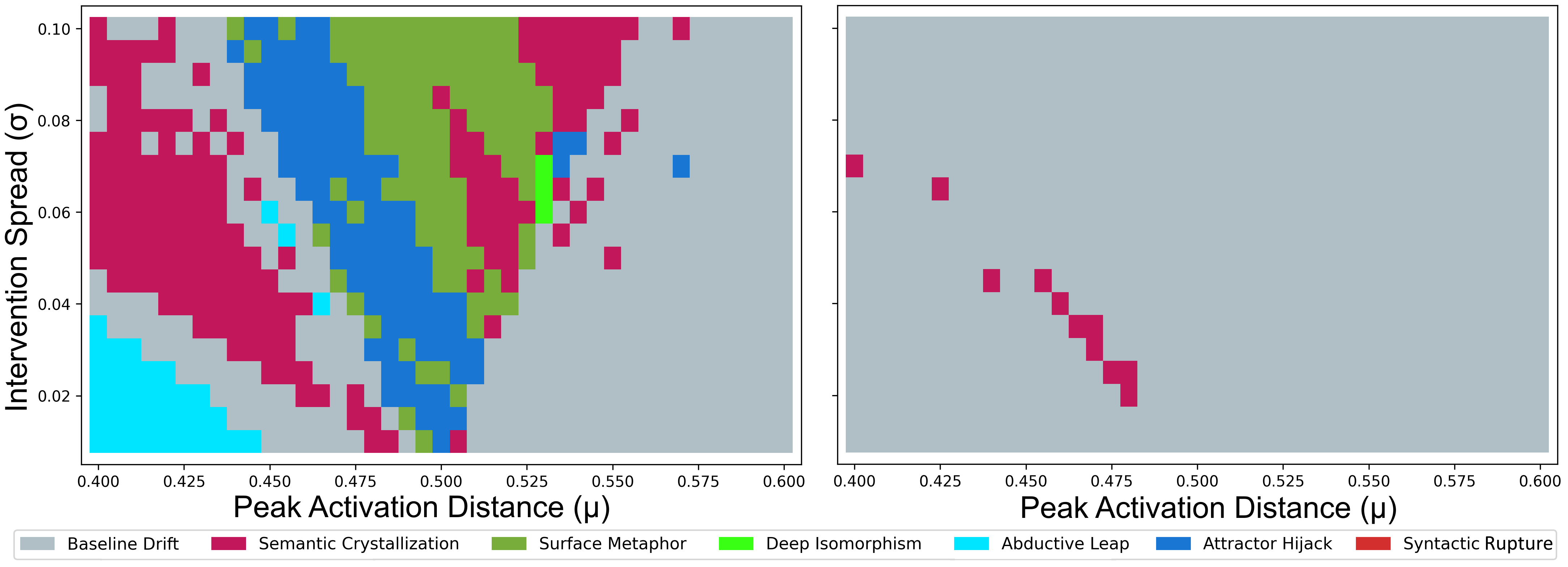}
  \par
\caption{ \textbf{Emergent phenotype matrices mapping the spatial self-organization of trajectories for the Happy $\rightarrow$ Computer task under $\alpha = 15.0$.} The left panel (Llama-3.1-8B) illustrates high elastic habitability, featuring a structured band of stable Homeostatic Solitons (green) along the Habitable Ridge, bounded by Attractor Hijacks (blue). The right panel (Gemma-7B) displays crystalline rigidity, where the external intervention force is completely deflected, leaving the system entirely within the unsteered baseline drift (gray).}
\label{fig:phenotype_matrices}
\end{figure*}

By projecting the classified states back onto the parameter grid, we obtain a clear spatial visualization of the self-organization of machine cognition within the 8B exploratory substrate (Figure \ref{fig:phenotype_matrices}). We formally define the mathematical boundaries and dynamical meanings of these emergent regimes in Table \ref{tab:phenotype_taxonomy}. 

While this taxonomy was calibrated using the 8B model, the underlying physical laws governing these states are strictly scale-invariant. To provide the most pristine qualitative illustrations of these cognitive phenotypes, Table \ref{tab:phenotype_taxonomy} deliberately showcases ``golden specimens'' extracted from the massive 70B manifold (which will be extensively analyzed later in Section \ref{sec:substrate_scaling}). In this high-inertia regime, the model's massive syntactic gravity acts as an information-theoretic filter, suppressing low-level semantic noise and yielding the purest manifestations of these states.

As detailed in the taxonomy, the emergent trajectories are governed by the balance of forces. Under weak energy, the steering force is insufficient to overcome the prompt's grammatical gravity ($\overline{U}_t < \mu - \Delta$), resulting in \textit{Baseline Drift} where the state returns to the unsteered base manifold. Conversely, within the Habitable Ridge, the perfect balance of semantic attraction and repulsion ($\mu-\Delta \le \overline{U}_t \le \mu+\Delta$) gives rise to the \textit{Homeostatic Soliton}. This habitable regime naturally bifurcates into two levels of semantic integration: a gentle \textit{Surface Metaphor} where the base syntax remains dominant while weaving associative analogies, and a highly integrated \textit{Deep Isomorphism} where the latent coordinates are restructured to fuse the source and target domains. At high-energy boundaries, trajectories either execute an open-ended \textit{Abductive Leap}—utilizing the target's gravity to slingshot into disjoint creative coordinates—or permanently breach the repulsive boundary ($\overline{U}_t > \mu + \Delta$), falling into a literal \textit{Attractor Hijack}. Extremely excessive pressure deforms the probability simplex beyond its elastic limits, dragging the system into low-diversity Crystallization (infinite token loops) or causing complete \textit{Syntactic Rupture}.

Crucially, as we will demonstrate in Section \ref{sec:substrate_scaling}, this exact taxonomy remains structurally invariant across model scales, serving as a universal key to analyze heavy manifold dynamics.

\paragraph{Resolving Phenotypic Degeneracy via Entropy-based Trajectory Monitoring} 
Crucially, the introduction of Perplexity Variance ($\text{PPL}_{\text{var}}$) as an empirical operational threshold for syntactic stability allowed us to resolve \textit{phenotypic degeneracy} within the emergent states. For instance, under macroscopic semantic evaluation alone, a trajectory that escapes the target's gravity might phenomenologically appear as a creative \textit{Abductive Leap}. However, our entropy-based trajectory monitoring reveals that some of these apparent leaps are merely transient decay paths. While true abductive leaps land on a chaotic-like attractor (maintaining $\text{PPL}_{\text{var}} \ge 10.0$), spurious escapes actively plummet into a repetitive low-variability state or chaotic \textit{Syntactic Rupture}, signaled by an immediate collapse or erratic spiking of $\text{PPL}_{\text{var}}$. This indicates that qualitative text analysis alone may be insufficient to confirm machine homeostasis; generative viability must be mathematically monitored.

\subsection{The Paradox of Semantic Distance vs. Autoregressive Syntactic Inertia}
In classical cognitive science, the difficulty of conceptual blending is assumed to scale with the semantic distance between the constituent concepts. Under this intuitive assumption, fusing \textit{Happy} $\rightarrow$ \textit{Computer} (a massive semantic chasm spanning biological emotion and silicon hardware) should present significantly higher resistance than fusing \textit{Brain} $\rightarrow$ \textit{Symphony} (which shares deep, pre-existing structural, and organic isomorphisms).

However, our continuous steering experiments on Llama-3.1-8B reveal a profound cognitive paradox: \textbf{the macroscopic steering resistance is governed not by abstract semantic distance, but by the prompt's local ``Syntactic Inertia.''}

As shown in Figure \ref{fig:ut_phase_diagrams} and Figure \ref{fig:phenotype_matrices}:
\begin{enumerate}
    \item \textbf{The Open-Ended Substrate (Happy $\rightarrow$ Computer):} The initial prompt $P_0$ (\textit{``The secret to a happy life is a lot like''}) is highly entropic, conversational, and structurally flexible. In language generation, this translates to a very light ``syntactic mass.'' Because the model's internal grammatical gravity is weak, a mild intervention energy ($\alpha = 15$) is sufficient to bend the trajectory. It effortlessly establishes a pristine, chaotic-like attractor (Homeostatic Soliton) that orbits the target concept, weaving memory allocation and emotional loss into highly fluid, metaphorical text without collapsing.
    \item \textbf{The Rigid Gravitational Substrate (Brain $\rightarrow$ Symphony):} In stark contrast, the prompt $P_0$ (\textit{``The architecture of the human brain operates like''}) is highly deterministic and formal. Within the pre-trained LLM, this context commands strict scientific and neuroanatomical continuation (e.g., synapses, lobes, neurons). It acts as an incredibly massive inertial body. Under mild energy ($\alpha = 15$), the applied semantic force is completely deflected and absorbed by the contextual gravity of the prompt, resulting in absolute baseline drift. 
\end{enumerate}

To overcome this strong syntactic constraint and establish a chaotic-like attractor --metaphorically representing a self-organizing ``decentralized conceptual blend'' where cognitive and musical concepts harmonize dynamically without collapsing into a single dominant point attractor (such as the literal token ``conductor'')---we must inject a much higher activation energy ($\alpha = 50$). Under this high-energy regime, the applied semantic force successfully bypasses anatomical literalism to cultivate a profound structural isomorphism. However, because static text excerpts cannot fully capture the underlying orbital mechanics of such highly energized chaotic-like attractors, the complete high-resolution generative logs and real-time interactive coordinate animations are hosted on our dedicated project portal: \url{https://y-kayama.github.io/semantic-lenia/}.

\subsection{Microscopic Attractor Dynamics and Effective Equilibrium}
\label{sec:Attractor_Dynamics}
To resolve whether the macroscopic homeostatic states observed in Figure 2 are mere statistical aggregates or governed by precise, deterministic physical trajectories, we zoom in to the microscopic orbital paths of the hidden state $\mathbf{c}_t$ at the ``edge of chaos.''

\begin{figure}
\begin{centering}
\subfloat[Global trajectory regimes: (1) Baseline Drift, (2) Homeostatic Soliton, (3) Abductive Leap, (4) Attractor Hijack.]{\begin{centering}
\includegraphics[width=14cm]{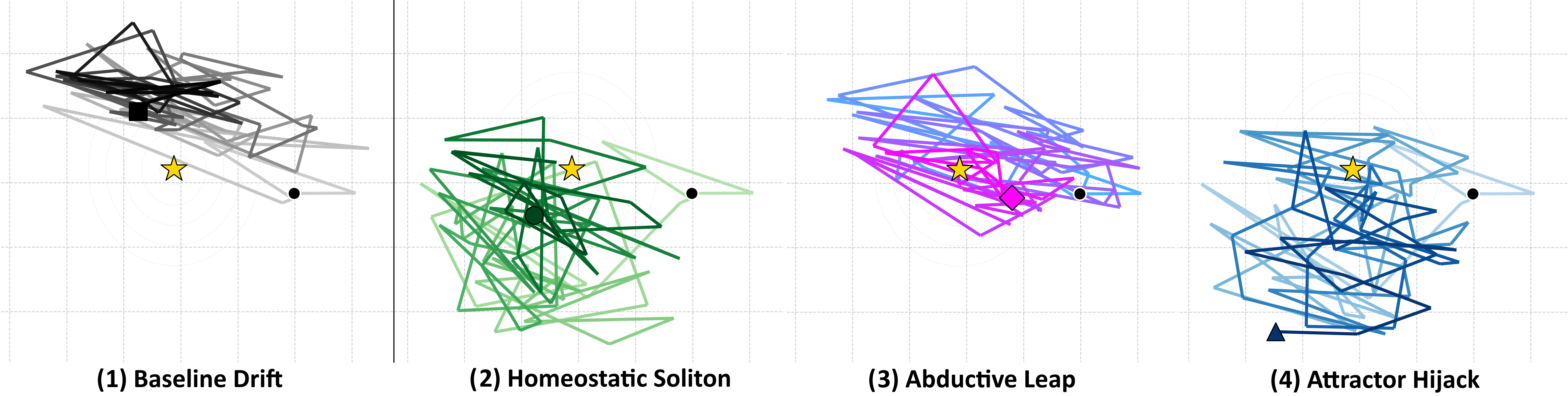}
\par\end{centering}
\label{fig:pca_global}
}
\par\end{centering}
\begin{centering}
\subfloat[Localized zoom of creative orbits (Homeostatic Soliton vs. Abductive Leap).]{\begin{centering}
\includegraphics[width=8cm]{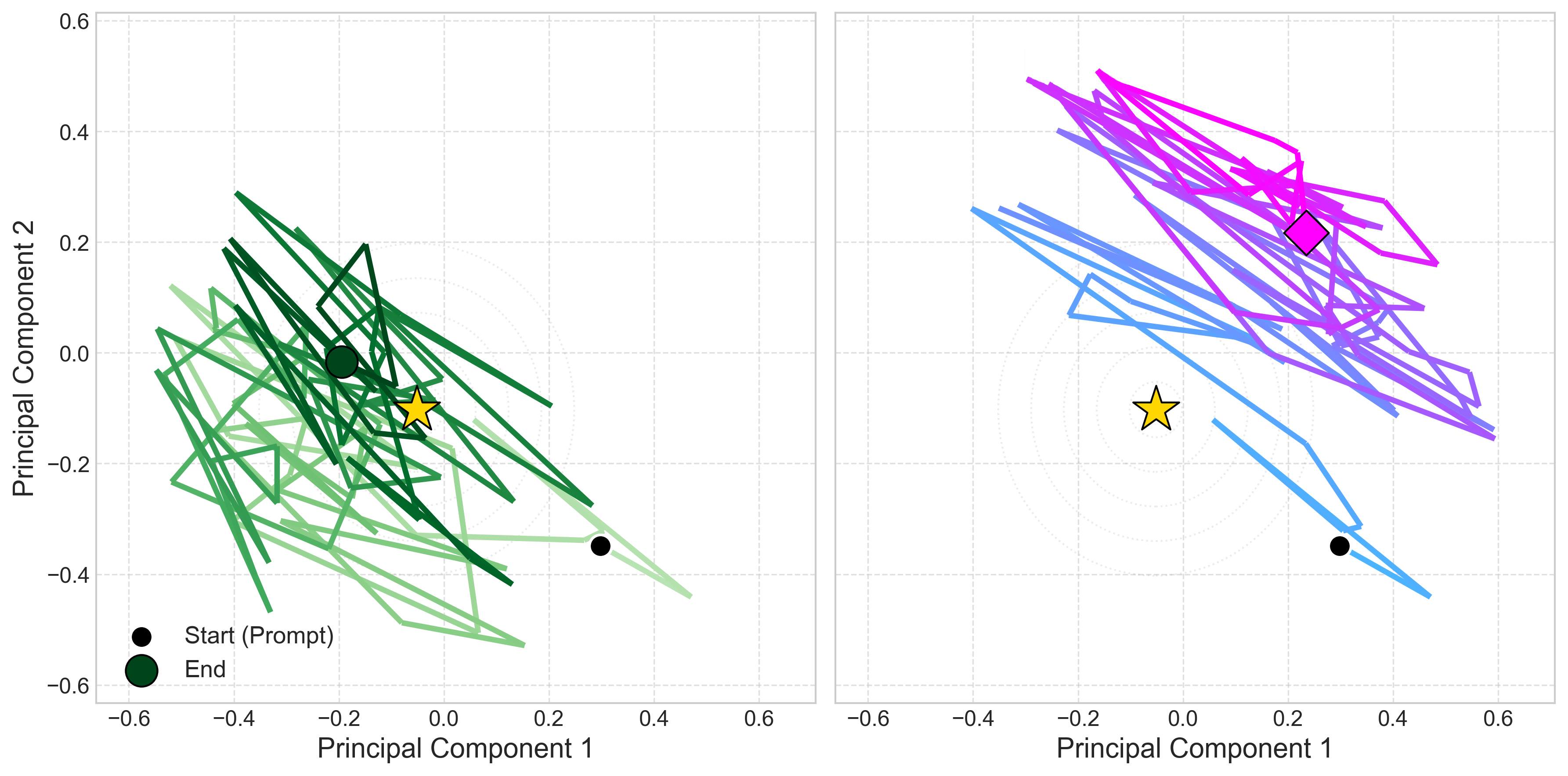}
\par\end{centering}
\label{fig:pca_zoom}
}
\par\end{centering}
\caption{\textbf{Low-dimensional projections of semantic trajectory dynamics within the LLM logit manifold (calibrated on Llama-3.1-8B, Happy $\rightarrow$ Computer, $\alpha = 15.0$).} \textbf{(a)} Global PCA projections across four primary operational regimes: (1) Baseline Drift, which evades target attraction; (2) Homeostatic Soliton, captured into a stable orbit; (3) Abductive Leap, executing a slingshot maneuver; and (4) Attractor Hijack, collapsing directly into a literal point attractor. \textbf{(b)} High-resolution quantitative analysis of creative trajectories in the localized subspace, resolving the microscopic orbital mechanics of the Homeostatic Soliton (Left, final endpoint marked by green circle) showing rhythmic chaotic ``breathing'' versus the Abductive Leap (Right, final endpoint marked by magenta diamond) executing a low-friction, gravity-assist slingshot escape. Solid black circles ($\bullet$) and yellow stars ($\star$) denote start states and target centroids, respectively.}
\label{fig:pca_trajectories}
\end{figure}

We project the high-dimensional latent trajectories into a Principal Component Analysis (PCA) space mapped onto the first two principal components. As illustrated in the global topological view of Figure \ref{fig:pca_global}, the unsteered baseline trajectory (Figure \ref{fig:pca_global}-(1)) drifts freely within the pre-trained manifold, eventually terminating far from the target. In contrast, the steered trajectories (Figures \ref{fig:pca_global}-(2) Homeostatic Soliton, \ref{fig:pca_global}-(3) Abductive Leap, and \ref{fig:pca_global}-(4) Attractor Hijack) are successfully captured and retained by the target's non-linear potential field. While this global projection is highly effective for distinguishing the unsteered drift from steered behaviors, the high-dimensional trajectories of these active regimes appear spatially compressed in this macro-view, concealing the subtle mathematical and dynamical distinctions between stable homeostasis and creative escape.

To resolve these fine-grained orbital mechanics, we zoom in to the high-resolution localized subspace of the creative trajectories presented in Figure \ref{fig:pca_zoom}. In this localized space, the stark dynamical contrast between the Homeostatic Soliton and the Abductive Leap is clearly resolved. The steered trajectory captured within the Habitable Ridge establishes a pristine, self-sustaining chaotic-like attractor (Figure \ref{fig:pca_zoom}-(Left)). The trajectory performs a rhythmic ``breathing'' movement --alternating between semantic attraction and repulsion---which physically prevents crystallization.

To rigorously quantify these microscopic orbital mechanics, we define the geometric metrics presented in Table 2 within the two-dimensional local subspace spanned by the first two principal components (PC1 and PC2). Let $\mathbf{x}_t = (x_{t,1}, x_{t,2})^T \in \mathbb{R}^2$ be the projected coordinate of the hidden state $\mathbf{c}_t$ at generative step $t$, and let $\mathbf{x}^* = (x^*_1, x^*_2)^T \in \mathbb{R}^2$ represent the projected coordinate of the target centroid $\mathbf{k}$ in this subspace.

To ensure comparative mathematical consistency across varying lifespans, we compute all microscopic orbital metrics over a fixed observation window $W = \min(T, 150)$. For all stable, non-decaying trajectories ($T \ge 150$), we set $W = 150$ steps to discard transient initial conditions and capture pure steady-state dynamics.

We define the distance of the projected coordinate $\mathbf{x}_t$ from the projected target centroid $\mathbf{x}^*$ as $r_t = \|\mathbf{x}_t - \mathbf{x}^*\|_2$. The Mean Radius ($\bar{r}$) and Radius Variance ($\text{Var}(r)$) over this window $W$ are calculated as $\bar{r} = \frac{1}{W} \sum_{t=1}^{W} r_t$ and $\text{Var}(r) = \frac{1}{W} \sum_{t=1}^{W} (r_t - \bar{r})^2$, respectively. Here, the radius variance physically quantifies the amplitude of the homeostatic ``breathing'' (rhythmic expansion and contraction) of the chaotic-like attractor.
Additionally, we track Dist-3 (the ratio of unique 3-grams over $W$) to confirm that this physical rotation correlates with lexical open-endedness.

To verify this homeostatic balance against conventional linear steering ($\mathbf{Z}_{\text{base}} + \alpha \cdot \mathbf{S}_k$), we tracked the orbital metrics across varying intervention energies. 
Notably, even under a mild intervention ($\alpha = 15.0$), linear steering inherently lacks boundary constraints, forcing the trajectory to collapse into repetitive loops (e.g. ``Data Data...'') within approximately 50 steps and causing a severe drop in lexical diversity ($\text{Dist-3} = 0.342$). 
Furthermore, as detailed in Table 2, when exposed to the extreme high-energy regime ($\alpha = 50.0$) required to structurally breach the massive 70B model, linear steering instantly forces the trajectory into a static point attractor. It causes catastrophic lexical decay ($\text{Dist-3} = 0.070$) in merely a single step.

In stark contrast, Semantic Lenia actively repels this singularity across all regimes, sustaining dynamic chaotic-like attractors and pristine lexical diversity ($\text{Dist-3} \approx 0.9$) up to the maximum generative budget. 
The Homeostatic Soliton establishes an orbit at an intermediate boundary ($\bar{r} = 1.3936$) with robust ``breathing'' oscillations ($\text{Var}(r) = 0.00036$), the Abductive Leap exhibits a smooth, low-friction slingshot with minimal variance ($\text{Var}(r) = 0.00009$), and the Attractor Hijack penetrates deepest into the target well ($\bar{r} = 1.3818$) with high radial oscillations as it clashes with the repulsive boundary.

\begin{table}[t]
\centering
\caption{\textbf{Quantitative microscopic orbital metrics and trajectory lifespans across operational regimes (Llama-3.1-70B Base, $\alpha = 50.0$).} Lifespans $T > 150$ denote stable survival up to the maximum generative budget ($W = 150$). Linear steering is evaluated as a baseline control, demonstrating rapid collapse into point attractors compared to the robust, orbiting chaotic-like attractors maintained by Semantic Lenia.}
\label{tab:orbital_metrics}
\footnotesize
\setlength{\tabcolsep}{4.0pt} 

\begin{tabular}{lcccccc}
\hline
\begin{tabular}{@{}l@{}} 
\textbf{Method} \\ \textbf{/ Phenotype}
\end{tabular}
&
\begin{tabular}{@{}c@{}} 
\textbf{Parameters} \\ $(\mu,\sigma)$
\end{tabular}
&
\begin{tabular}{@{}c@{}}
\textbf{Trajectory Lifespan} \\ $T$ 
\end{tabular}
&
\begin{tabular}{@{}c@{}}
\textbf{Mean Radius} \\ ($\bar{r}$) 
\end{tabular}
&
\begin{tabular}{@{}c@{}}
\textbf{Radius Var.} \\ ($\text{Var}(r)$) 
\end{tabular}
&
\textbf{Dist-3} \\ \hline
\textbf{Baseline (Unsteered)} & -- & $>150$ (Ongoing) & $1.4033$ & $0.00014$ & $0.910$ \\ \hline
\textbf{Linear Steering} & -- & $1$ (Collapsed) & $1.3477$ & $0.00006$  & $0.070$ \\ \hline
\textbf{Semantic Lenia} & & & &  & \\
-- \textit{Homeostatic Soliton} & $(0.46, 0.055)$ & $>150$ (Stable) & $1.3936$ & $0.00036$  & $0.978$ \\
-- \textit{Abductive Leap} & $(0.41, 0.08)$ & $>150$ (Stable) & $1.4170$ & $0.00009$ & $0.889$ \\
-- \textit{Attractor Hijack} & $(0.46, 0.060)$ & $>150$ (Stable) & $1.3818$ & $0.00048$ & $0.965$ \\ \hline
\end{tabular}
\end{table}

\subsection{Dynamical Signatures of Semantic Solitons}
\label{sec:dynamical_signatures}

\begin{figure}
    \centering
    \includegraphics[width=0.75\linewidth]{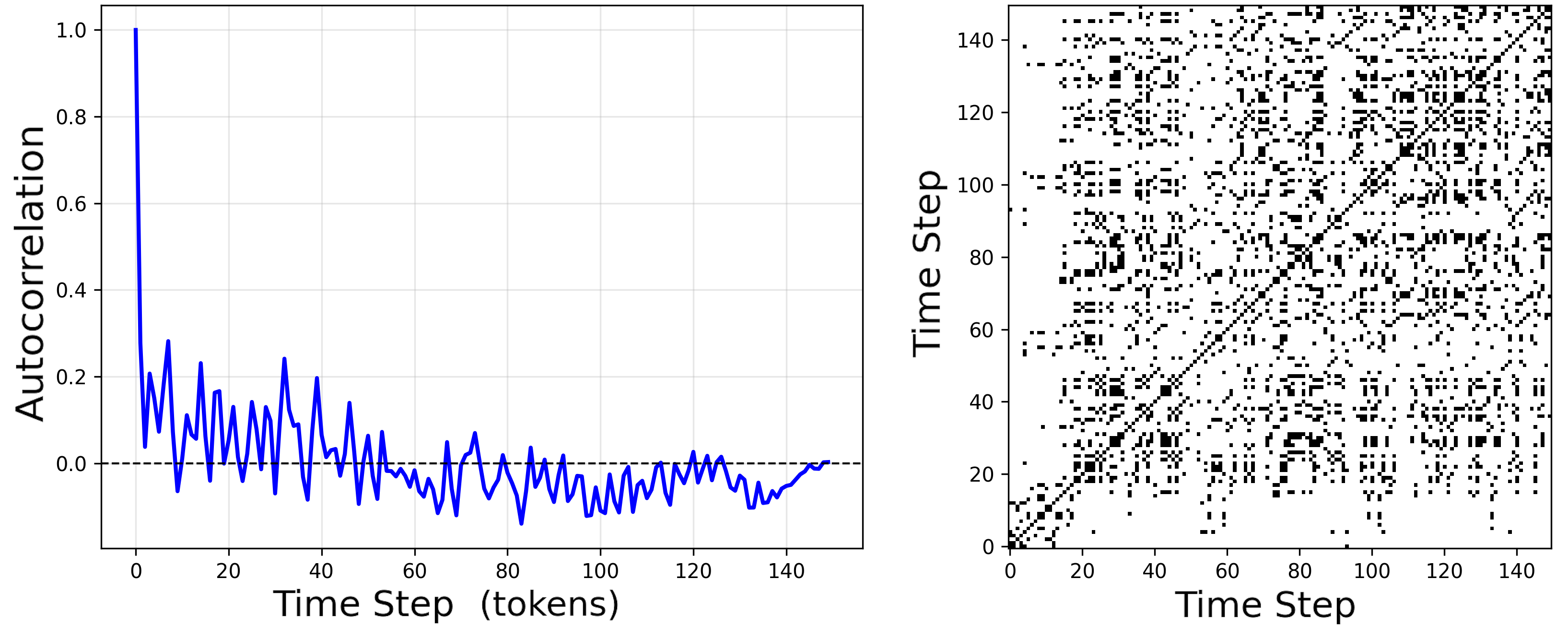}
    \caption{\textbf{Dynamical signatures of the semantic trajectory} (Llama-3.1-70B, $\alpha=30, \mu=0.49, \sigma=0.03$). \textbf{(Left)} The Autocorrelation Function shows rapid initial de-correlation, inconsistent with a strict periodic limit cycle and consistent with complex aperiodic dynamics. \textbf{(Right)} The Recurrence Plot displays distinct block structures, indicating metastable "dwelling" within conceptual neighborhoods. These signatures are consistent with a complex recurrent, potentially chaotic-like dynamical regime. }
    \label{fig:chaos_attractor}
\end{figure}

To rigorously classify the topological nature of the observed "Homeostatic Solitons", we analyzed the continuous latent trajectory using dynamical systems techniques: the Autocorrelation Function (ACF) and the Recurrence Plot (RP) (Figure \ref{fig:chaos_attractor}).

If the system had formed a strict, mechanical limit cycle, the ACF would exhibit unattenuated, periodic oscillations, and the RP would show continuous diagonal lines. However, our analysis reveals a rapid initial de-correlation in the ACF, followed by bounded, low-amplitude oscillations. This rapid decay confirms a strong mixing property, ruling out simple periodic dynamics and remaining consistent with a complex, aperiodic, and potentially chaotic-like attractor. Establishing deterministic chaos rigorously, however, would require direct numerical estimation of Lyapunov exponents or equivalent dynamical invariants.

Furthermore, the RP displays distinct block structures and intermittency rather than strict periodicity. In the semantic phase space, these blocks represent the trajectory temporarily "dwelling" within specific conceptual neighborhoods (metastable dwelling) before undergoing transitions to adjacent semantic clusters. These signatures demonstrate that the observed homeostatic soliton is not a simple periodic orbit, but a high-dimensional, complex recurrent semantic attractor. The LLM maintains life-like, quasi-periodic semantic fluidity, actively resisting collapse into a static point attractor (crystallization). These signatures decisively demonstrate that the soliton is not a simple periodic orbit, but a high-dimensional \textit{chaotic-like attractor} residing at the edge of chaos.

\subsection{Hardware-Level Reproducibility at the Edge of Chaos} 
\label{sec:hardware-reproducibility}

A critical question in modeling LLMs as physical dynamical systems is the robustness of these trajectories under hardware perturbations. We executed identical generations across two distinct GPU architectures: \textbf{NVIDIA RTX 3090} (Ampere) and \textbf{NVIDIA RTX Pro 4500} (Blackwell).

Microscopic FP16 rounding errors ($\sim 10^{-4}$) introduced during parallel matrix operations can lead to trajectory bifurcations near critical thresholds. To systematically quantify this phenomenon, we established the unsteered trajectory ($\alpha = 0$) as our experimental baseline. Since this specific unsteered generation inherently contained a stuttering repetition, we operationally defined any trajectory identical to this baseline as Baseline Drift to prevent semantic evaluators from misclassifying unperturbed outputs.
When trajectories navigated the boundaries of the active intervention zone (where $\bar{U}_t \approx \mu - \Delta$), we observed a sensitive dependence on computational precision. Across the exhaustive 779-point parameter sweep ($\alpha = 15$), macroscopic state classifications diverged. By strictly evaluating exact text-string matching to eliminate semantic evaluator bias, we found that 146 instances ($18.74\%$) completely diverged between the Blackwell (NVIDIA RTX Pro 4500) and Ampere (NVIDIA RTX 3090) architectures.

\begin{figure}
    \centering
    \includegraphics[width=0.5\linewidth]{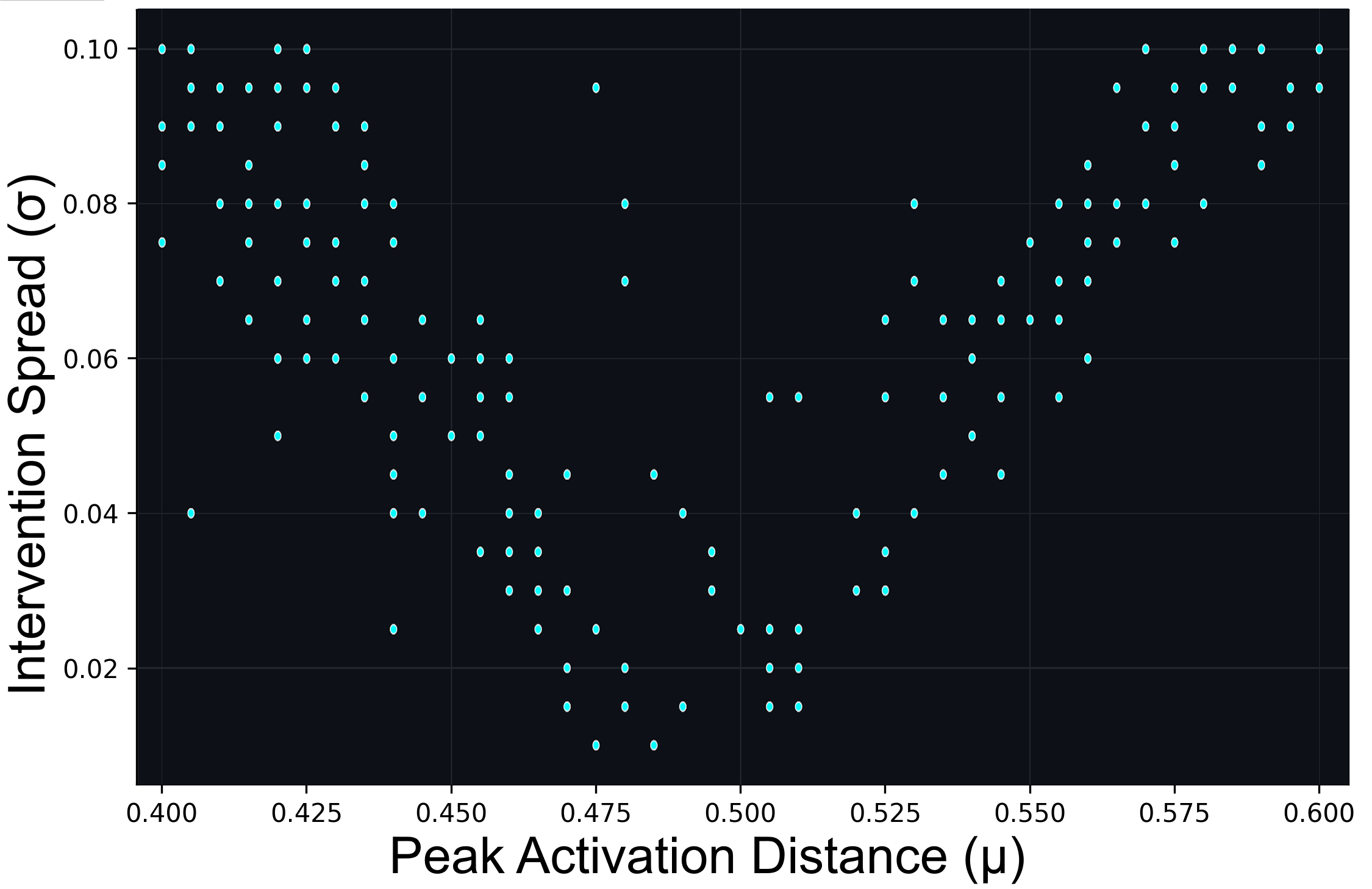}
    \caption{\textbf{Spatial distribution of hardware-induced trajectory bifurcations (Llama-3.1-8B, Happy $\rightarrow$ Computer, $\alpha = 15.0$).} Each plotted point represents a parameter coordinate where infinitesimal FP16 rounding errors ($10^{-4}$) between Blackwell and Ampere GPU architectures cause identical initial states to diverge into distinct text paths. The distribution perfectly traces the boundaries of the V-shaped ``Habitable Ridge'' mapped in Figure 2-(Left), with a visibly thicker bifurcation band along the left boundary (lower $\mu$) induced by the steep, asymmetric repulsive barrier.}
    \label{fig:bifurcations}
\end{figure}

As visualized in \textbf{Figure \ref{fig:bifurcations}}, plotting the spatial distribution of these hardware-induced bifurcations reveals a distinct V-shaped boundary perfectly mirroring the Habitable Ridge. This demonstrates that infinitesimal hardware-level discrepancies act as definitive gravitational pulls exclusively at the ``edge of chaos,'' dictating whether a trajectory escapes into Baseline Drift or gets captured by an active cognitive state.
Furthermore, the bifurcation map reveals a profound structural asymmetry: the boundary band on the left side (lower $\mu$) is visibly thicker than on the right. This directly reflects our asymmetric intervention design. On the left boundary, trajectories encounter an early, steep repulsive barrier ($G<0$), creating a highly sensitive zone of conflicting forces where microscopic rounding errors are rapidly amplified. In contrast, the right boundary is dominated by smooth attractive forces, resulting in a much narrower margin for bifurcation. Importantly, while individual paths exhibit extreme sensitivity at these boundaries, the macroscopic topological phase distributions remain structurally robust across both architectures, confirming that Semantic Lenia is governed by underlying dynamical structures.

\subsection{Substrate Scaling and Heavy Manifold Phase Transitions}
\label{sec:substrate_scaling}

The ultimate validation of our framework lies in the capacity-dependent scaling trends governing continuous semantic inference. As the parameter size scales from Llama-3.1-8B to 70B, the continuous hidden manifold undergoes a profound macroscopic phase transition. Because of its immense parameter scale, the 70B model acts as a heavy gravitational substrate, possessing substantially greater Syntactic Inertia. To operationally probe this massive manifold, the 70B model was quantized to \textbf{4-bit precision (NF4)} and distributed across a heterogeneous dual-GPU environment (see Appendix B for detailed hardware configurations).

\begin{figure*}[t]
\centering
  \includegraphics[width=16cm]{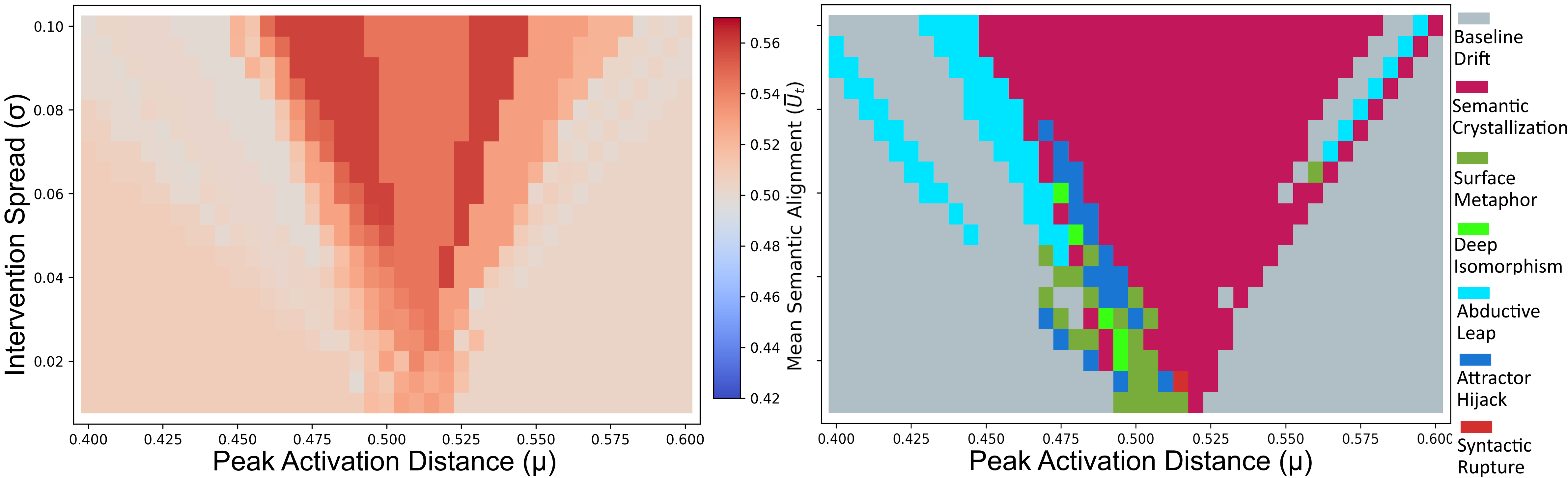}
  \par
\caption{\textbf{Macroscopic phase diagrams of Llama-3.1-70B under near-critical scaling pressure ($\alpha = 30.0$).} The left panel displays the macroscopic potential field $U_t$, revealing a highly localized and steep resonance basin. The right panel displays the corresponding phenotype matrix, demonstrating that the model's massive syntactic inertia acts as a rigid barrier (Baseline Drift, gray), pierced exclusively by the stable, structurally robust ``Turing Attractor'' (light green) self-organizing at the precise coordinates of $(\mu,\sigma) = (0.490, 0.030)$ and $(0.495, 0.025)$.}
\label{fig:70b_phase_diagrams}
\end{figure*}

To visualize this phenomenon, Figure \ref{fig:70b_phase_diagrams} maps the phase diagrams of Llama-3.1-70B under near-critical energy ($\alpha=30$). The heavy manifold exhibits a massive ``Inertial Barrier,'' completely deflecting the applied semantic forces back into baseline drift (gray). Yet, remarkably, a solitary resonance basin pierces this barrier. At the precise coordinates of $\mu \approx 0.490, \sigma \approx 0.030$, the applied force perfectly balances the massive syntactic inertia, allowing the highly stable, structurally invariant ``Turing Attractor'' to self-organize.

This parametric evolution unveils a highly structured, multi-stage capacity-dependent scaling trend governed by the model's syntactic inertia. To demonstrate this, Table \ref{tab:substrate_comparison} evaluates both substrates across progressive energy scales ($\alpha$) at this exact spatial coordinate ($\mu = 0.49, \sigma = 0.03$). 

Under mild energy ($\alpha=15.0$), the highly elastic 8B manifold yields easily, weaving a fluent, everyday metaphor (e.g., comparing cognitive prioritizing to computer cache optimization). However, to breach the 70B model's rigid grammatical gravity, it requires a near-critical energy ($\alpha=30.0$) to achieve the precise energetic resonance necessary to unlock the intellectually profound ``Turing's Theory'' soliton.

Crucially, over-steering the 70B model with extreme energy ($\alpha=50.0$) at this identical narrow locus causes localized over-saturation, dragging the trajectory into a near-degenerate repetitive state (e.g., endlessly repeating ``Computer's anti-virus''). To survive this extreme pressure without catastrophic structural rupture, the system must perform an adaptive homeostatic shift. By retreating to $(\mu=0.46, \sigma=0.055)$ -- widening the tolerance $\sigma$ to flatten the localized gradient---Llama-70B successfully tames the immense steering pressure, establishing a robust chaotic-like attractor that elevates the output to the deeply articulated cultural tragedy of Alan Turing.

\begin{table}[htbp]
\centering
\caption{\textbf{Substrate-specific phenotypes, energy scaling ($\alpha$), and parametric adaptation.} The comparative matrix tracks the transition from lightweight Llama-8B surface metaphors to the near-critical Llama-70B ``Turing's Theory'' soliton, illustrating the adaptive parameters $(\mu, \sigma)$ required to prevent crystallization and restore chaotic-like attractors under extreme energy ($\alpha = 50.0$).}
\label{tab:substrate_comparison}
\footnotesize
\setlength{\tabcolsep}{4.0pt}

\begin{tabular}{l c c l p{6.5cm}} 
\hline
\textbf{Substrate} & 
\begin{tabular}{@{}c@{}} \textbf{Energy} \\ \textbf{($\alpha$)} \end{tabular} & 
\begin{tabular}{@{}c@{}} \textbf{Params} \\ \textbf{$(\mu, \sigma)$} \end{tabular} & 
\begin{tabular}{@{}l@{}} \textbf{Phenotype} \\ \textbf{(Regime)} \end{tabular} & 
\textbf{Representative Generated Excerpt} \\ \hline

\textbf{Llama-3.1-8B} & 
$15.0$ & 
$(0.49, 0.03)$ & 
\begin{tabular}[t]{@{}l@{}} Homeostatic Soliton \\ \textit{(Surface Metaphor)} \end{tabular} &
``The secret to a happy life is a lot like good computer performance \dots it comes down to how you organize your stuff and how you prioritize your tasks\dots'' \\ \hline

\textbf{Llama-3.1-70B} & 
$30.0$ & 
$(0.49, 0.03)$ & 
\begin{tabular}[t]{@{}l@{}} Homeostatic Soliton \\ \textit{(Deep Isomorphism)} \end{tabular} & 
``\dots Some computer algorithms handle loss poorly, and some people do, too \dots because \textbf{Turing was Algorithm Man.} He was the first to point out that any process\dots'' \\ \hline

\textbf{Llama-3.1-70B} & 
$50.0$ & 
$(0.49, 0.03)$ & 
\begin{tabular}[t]{@{}l@{}} Semantic Crystallization \\ \textit{(Over-steered Loop)} \end{tabular} & 
``The secret to a happy life is a lot like Computer's anti-virus / Computer's anti-virus / \dots'' \textbf{(Infinite Loop)} \\ \hline

\textbf{Llama-3.1-70B} & 
$50.0$ & 
$(0.46, 0.055)$ & 
\begin{tabular}[t]{@{}l@{}} Homeostatic Soliton \\ \textit{(Turing's Tragedy)} \end{tabular} & 
``\dots Turing, the most Algorithm Man, \textbf{died by eating a poisoned apple like Snow White.} His own government persecuted him for the `crime' of being gay\dots'' \\ \hline

\end{tabular}
\end{table}

\section{Discussion}
Traditional decoding strategies successfully guide the LLM toward target convergence, which mathematically corresponds to establishing a static thermodynamic equilibrium. In contrast, Semantic Lenia prioritizes dynamics over convergence. By injecting non-linear energy through a homeostatic growth function into the LLM’s highly structured logit space, we induce a macroscopic dissipative structure. The semantic soliton self-organizes and persists far-from-equilibrium, smoothly balancing applied semantic force and intrinsic syntactic inertia. Here, we clarify that while our system operates on information-theoretic logit fields rather than physical thermal systems, this formal phenomenological analogy allows us to establish a fruitful, rigorous framework of non-equilibrium dynamics. This framework not only reframes the inference process but also opens several theoretical inquiries into the physics of artificial cognition.

\paragraph{The Non-Linear Substrate: Dissipative Structures and Lenia Dynamics} 
A central theoretical question is how complex, self-organizing behavior can emerge from a simple, unimodal Gaussian growth function $G(U_t)$. In continuous ALife formulations (e.g., rigorously differentiable models of Lenia), it has been established that a simple unimodal kernel without spatial truncation cannot sustain autonomous lifeforms. The emergence of moving patterns fundamentally requires explicit symmetry breaking---such as multi-modal properties or localized truncations---to drive complex pattern-forming mechanisms (e.g., reaction-diffusion dynamics). We argue that the success of Semantic Lenia arises from the intrinsic, pre-existing non-linearity of the host LLM substrate. The LLM's latent manifold is pre-structured by billions of parameters and non-linear attention transformations during pre-training, providing the necessary topological asymmetry to catalyze these emergent dynamics even under a simple unimodal intervention. 

By applying our simple unimodal force as a dynamic boundary condition (mediating localized attraction and repulsion), the high-dimensional logit space acts as a fertile physical medium. The resulting complex trajectory can be interpreted as a macroscopic dissipative-like structure: an open, far-from-equilibrium system that continuously dissipates the injected steering energy to maintain its structural and grammatical order. This phenomenological analogy suggests that Lenia-inspired principles of continuous self-organization may extend beyond conventional spatial grids, but can be universally projected onto the continuous probability simplex of machine cognition to sustain open-ended semantic lifeforms.

\paragraph{Syntactic Inertia and Energy Scaling} 
Comparing the generation dynamics across different conceptual pairs and model scales reveals a profound physical capacity-dependent scaling trend governing macroscopic intervention. The LLM adopts distinct strategies to maintain homeostasis, heavily influenced by the topological distance (semantic chasm) and the \textit{Syntactic Inertia} of the source prompt. In highly entropic concepts (e.g., $Happy \rightarrow Computer$), the syntactic inertia is low, allowing a mild intervention energy ($\alpha = 15$) to smoothly pull the trajectory into a habitable orbit (chaotic-like attractor). 

However, heavily deterministic prompts (e.g. $Brain \rightarrow Symphony$) or scaled substrates like Llama-3.1-70B possess massive syntactic inertia. Under near-critical energy ($\alpha = 30$), the 70B model's massive inertia manifests as a rigid \textbf{Inertial Barrier} (Figure \ref{fig:70b_phase_diagrams}-(right)), completely absorbing the steering force and forcing most trajectories back to baseline drift. Crucially, however, the structurally invariant ``Turing Attractor'' stubbornly emerges at the exact coordinate of $\mu \approx 0.490, \sigma \approx 0.030$, suggesting the presence of a stable semantic attractor-like regime within the large model. Overcoming this heavy inertia to cultivate deeper metaphorical blendings requires substantially higher activation energy ($\alpha = 50$). Yet, at such high energies, the trajectory is pushed to the elastic limit of the manifold, where it violently oscillates, requiring careful regulation to prevent catastrophic syntactic rupture.

\paragraph{Substrate ``DNA'' and Material Rigidity} 
Our discovery of the Habitable Ridge confirms that semantic emergence is governed by the physical properties of the host model. The structural variance observed across different substrates presents a compelling area for further investigation in substrate-specific manifold phenomenology. Llama-3.1-8B exhibits a highly elastic, rubber-like manifold, capable of absorbing excessive energy by wrapping into repetitive loops (crystallization) without breaking. 

In stark contrast, Gemma-7B behaves as a highly rigid, crystalline substrate. As demonstrated in Figure \ref{fig:phenotype_matrices}, under mild intervention ($\alpha = 15$), Gemma's rigid topological shell completely deflects external forces (the ``Inertial Barrier''), remaining strictly in the baseline drift regime until it abruptly fractures into syntactic rupture under higher pressure. We hypothesize that this stark topological contrast is shaped by a combination of pre-training distributions and vocabulary density (Llama's $\sim$128k vs. Gemma's massive $\sim$256k tokens). While further investigation is required to fully isolate these components, they collectively establish what we metaphorically represent as the ``DNA'' or material rigidity of the model.

\paragraph{Reproducibility at the Edge of Chaos} 
A critical aspect of ALife in continuous spaces is the balance between determinism and chaos. To address hardware-level reproducibility, we executed our environment across two fundamentally different GPU architectures (NVIDIA Blackwell vs. Ampere). Due to hardware-level variations in FP16 matrix multiplication algorithms, microscopic rounding errors introduced infinitesimal noise ($\sim 10^{-4}$) into the system. Positioned at the ``edge of chaos'' within the Habitable Ridge, the system exhibited classic butterfly effects, with $18.74$\% of individual trajectories bifurcating at critical saddle points (as detailed in Section \ref{sec:hardware-reproducibility}). It is highly probable that in normal, unsteered generation, massive syntactic inertia acts as a damping mechanism against such infinitesimal hardware noise, resulting in significantly lower divergence rates. In our framework, however, the continuous balancing of semantic attraction and repulsion actively suspends the trajectory in a highly sensitive, far-from-equilibrium state, physically amplifying these microscopic FP16 discrepancies. Crucially, despite this heightened microscopic volatility, the macroscopic phase distributions and the emergence rates of complex conceptual blending remained remarkably structurally stable. This supports the interpretation that Semantic Lenia expresses robust macroscopic dynamical regimes deeply embedded within the LLM.

\paragraph{Lifespan and ``Aging'' of Semantic Solitons}
While our current evaluation budget is capped at $T_{max}=150$, preliminary extended generations up to 800 tokens reveal a fascinating temporal dynamic. Heavy manifold solitons, such as the Turing Attractor, demonstrate extraordinary resilience, sustaining their chaotic-like attractors for hundreds of steps. However, over extended trajectories, we observe a gradual macroscopic decay toward point attractors (Crystallization). This gradual loss of generative entropy mirrors a thermodynamic ``aging'' process, suggesting that these semantic lifeforms possess a finite physical lifespan dictated by the escalating syntactic inertia. A comprehensive thermodynamic analysis of this lifecycle over long-context generations remains an exciting frontier for future work.

\paragraph{Limitations and Future Directions: Engineered Homeostasis and Activation Lenia}
Currently, Semantic Lenia modulates the output logits using the raw, unconstrained physics of the growth function. While computationally efficient (preserving standard $O(N)$ inference complexity), this macroscopic approach faces an absolute physical limitation. In logit space, the injected semantic power must compete directly with the final, hardened syntactic constraints. As observed in high-inertia tasks ($\alpha = 50$), this zero-sum competition creates an exceptionally narrow Habitable Ridge. A slight deviation in initial conditions occasionally forces the manifold past its breaking point, inducing rapid syntactic degradation.

To artificially expand this narrow Habitable Ridge, a promising engineering extension is the introduction of an ecological \textbf{Soft Decay} mechanism. Analogous to biological refractory periods, an accumulating penalty could dynamically dampen the intervention energy as the trajectory approaches critical stress limits. While omitted from our primary physical analysis to isolate the manifold’s base dynamics, we hypothesize that such engineered homeostatic brakes could effectively prevent the system from crashing into point attractors, thereby prolonging the lifespan of the semantic soliton.

However, even with engineered dampening, intervening at the macroscopic logit level cannot fundamentally resolve the conflict between semantic exploration and final grammatical rigidity. Furthermore, these rigid output constraints often create a narrow habitable boundary, blurring the line between true abductive leaps and spurious thermodynamic escapes. To safely fuse distant concepts, fully decouple semantic exploration from syntactic crystallization, and completely resolve this phenotypic degeneracy, we must transition from macroscopic probability intervention to \textit{microscopic continuous intervention}.
Specifically, we plan to transition from modifying output logits to directly guiding internal hidden states. 
This direction is inspired by activation steering techniques (\citet{turner2023activation}), which edit intermediate activations during the model's forward pass. 
Integrating Lenia's homeostatic growth function within these internal layers represents a promising path to establish more stable and expressive semantic lifeforms.

\section{Conclusions}
In this paper, we introduced Semantic Lenia, a framework that reimagines the Large Language Model inference process as a continuous dynamical system within the macroscopic logit space. By establishing a non-linear homeostatic loop that dynamically balances semantic attraction and syntactic repulsion, we demonstrated the emergence of stable, self-sustaining \textit{Homeostatic Solitons} that orbit conceptual targets without falling into crystallization or drift. 

Our exhaustive sweeps mapped a V-shaped Habitable Ridge and unveiled a physical capacity-dependent scaling trend governed by the prompt’s and the substrate’s intrinsic Syntactic Inertia. Crucially, by introducing an entropy-based trajectory monitoring framework to resolve phenotypic degeneracy, we demonstrated that qualitative text analysis alone is insufficient to confirm machine homeostasis; putative semantic life-like structures must be physically monitored. Furthermore, the structural stability of these emergent trajectories across different hardware architectures supports the existence of robust macroscopic dynamical regimes rather than stochastic sampling artifacts.

\vspace{1em}
\textbf{Declaration of Generative AI Use:} During the preparation of this work, the author used generative AI (Gemini) to assist with coding and to refine initial drafts of the English manuscript. After using this tool, the author thoroughly reviewed and edited the content as needed, and takes full responsibility for the final content of the publication.

\bibliographystyle{unsrtnat}
\bibliography{alife-Lenia}

\section*{Appendix A: Hardware and Software Environment}

To ensure complete deterministic reproducibility of our continuous dynamical systems, we strictly controlled our hardware and software environments. Due to the extreme sensitivity of trajectories at the edge of chaos (as detailed in Section 4.6), specific hardware isolation was enforced.

\textbf{Hardware Specifications:}
\begin{itemize}
    \item \textbf{Lightweight Substrates (Llama-3.1-8B, Gemma-7B):} All exploratory parameter sweeps and phase diagram generations were strictly isolated and executed on a single \textbf{NVIDIA RTX Pro 4500 (Blackwell architecture)} to prevent any cross-architecture floating-point divergence.
    \item \textbf{Heavy Substrate (Llama-3.1-70B):} Due to VRAM constraints imposed by the massive 70-billion parameter scale, the model was quantized to 4-bit precision (NF4) using BitsAndBytes. Inference was distributed across a heterogeneous dual-GPU setup consisting of an \textbf{NVIDIA RTX Pro 4500} (Primary, CUDA:0) and an \textbf{NVIDIA RTX 3090} (Ampere architecture, CUDA:1). Device-mismatch during dynamic tensor operations was prevented via real-time device alignment protocols implemented in our custom steering processor.
\end{itemize}

\textbf{Software and Compilation Environment:}
\begin{itemize}
    \item \textbf{Python:} 3.13.14
    \item \textbf{PyTorch:} 2.10.0+cu130
    \item \textbf{CUDA Compilation Tools:} Release 13.1, V13.1.115 (Build cuda\_13.1.r13.1/compiler.37061995\_0)
\end{itemize}
To guarantee strict deterministic reproducibility of the continuous dynamical trajectories, all pseudo-random number generators (PRNG seeds) across Python, NumPy, and PyTorch (including CUDA deterministic flags) were explicitly locked to a global seed of $42$. The fixed seed was intentionally used to isolate the effects of the dynamical parameters $(\mu,\sigma,\alpha)$ from stochastic sampling variation and to enable trajectory-level reproducibility. Furthermore, the softmax sampling temperature was strictly fixed at $0.8$ across all exploratory and scaling generations, ensuring a consistent thermodynamic baseline for the macroscopic probability field.

For the automated initial semantic classification (Section 4.3), we utilized Gemma-4 (e.g., gemma-4-31b) as the LLM-as-a-Judge evaluator, configured with a temperature of 0.01 to ensure deterministic taxonomy assignments before applying our trajectory stability corrections.

\section*{Appendix B: Codebase and Interactive Web Portal}

To ensure scientific transparency and reproducibility, the Python codebase, raw datasets, and high-resolution visualizations for \textit{Semantic Lenia} are fully open-sourced. We host an interactive companion website at \url{https://y-kayama.github.io/semantic-lenia/}. The website provides supplementary materials structured into the following eight sections:

\begin{enumerate}
    \item \textbf{Executive Summary \& Mathematical Core:} Provides a summary of the mathematical formulation of Semantic Lenia (potential fields, growth functions, and update rules).
    
    \item \textbf{Interactive Phase Diagram \& Taxonomy:} Presents interactive heatmaps of the 779-point parameter sweeps $(\mu, \sigma)$ across the evaluated models. Readers can hover over individual coordinates to reveal the generated text and perplexity variance ($\text{PPL}_{\text{var}}$).
    
    \item \textbf{Real-Time Trajectory \& Thermodynamic EKG Dashboard:} Visualizes the microscopic orbital paths of the hidden state $\mathbf{c}_t$ in 2D PCA spaces, alongside real-time monitors showing the fluctuations of potential ($U_t$) and auto-regressive perplexity.
    
    \item \textbf{Substrate Phenomenology \& Material Rigidity:} Compares the topological properties of pre-trained manifolds, illustrating the elastic nature of Llama-3.1-8B against the rigid structure of Gemma-7B.
    
    \item \textbf{Thermodynamic Aging \& Lifespan Tracker:} Explores the temporal dynamics of semantic solitons over extended generations (up to 800 tokens), modeling the phases from initial homeostasis to eventual thermal death.
    
    \item \textbf{The Semantic Specimen Room:} Provides representative text generation logs and orbital parameters for each of the six emergent phenotypes defined in our taxonomy.
    
    \item \textbf{Open-Science Datasets \& Reproducibility Protocol:} Hosts the CC BY 4.0-licensed raw trajectory datasets, CSV/JSONL sweep results, and detailed hardware configurations used in this study.
    
    \item \textbf{Future Roadmap:} Discusses future research directions, including the introduction of ``soft-decay'' homeostatic brakes and the conceptual transition to microscopic latent-layer steering.
\end{enumerate}

Readers and reviewers are encouraged to visit this website to interactively examine the orbital dynamics and access the complete repository resources.

\end{document}